\documentclass{article}

 \usepackage[preprint]{neurips_2025}

\usepackage[utf8]{inputenc} 
\usepackage[T1]{fontenc}    
\usepackage{hyperref}       
\usepackage{url}            
\usepackage{booktabs}       
\usepackage{amsfonts}       
\usepackage{nicefrac}       
\usepackage{microtype}      
\usepackage{xcolor}         
\usepackage{graphicx}
\usepackage{subcaption}

\title{In Defence of Post-hoc Explainability}

\author{%
  Nick Oh \\
  socius labs \\
  \texttt{nick.sh.oh@socius.org} \\
}

\begin{document}

\maketitle

\begin{abstract}
This position paper defends post-hoc explainability methods as legitimate tools for scientific knowledge production in machine learning. Addressing criticism of these methods' reliability and epistemic status, we develop a philosophical framework grounded in mediated understanding and bounded factivity. We argue that scientific insights can emerge through structured interpretation of model behaviour without requiring complete mechanistic transparency, provided explanations acknowledge their approximative nature and undergo rigorous empirical validation. Through analysis of recent biomedical ML applications, we demonstrate how post-hoc methods, when properly integrated into scientific practice, generate novel hypotheses and advance phenomenal understanding.
\end{abstract}

\section{Introduction}

\textit{Is explainable AI (XAI) in its peril} \citep{weber2024xai}? Perhaps. However, this view has yet to gain mainstream acceptance. Driven by the increasing complexity and black-box nature of machine learning models, the field has witnessed an unprecedented surge in post-hoc explanation methods since 2017 \citep{saeed2023explainable, nagahisarchoghaei2023empirical, retzlaff2024post}, with bibliometric analyses documenting exponential growth in XAI approaches \citep{sharma2024exploring}. This momentum manifests in substantial industry adoption exemplified by tools such as \texttt{Captum} \citep{kokhlikyan2020captum}. 

However, this popularity-driven surge cannot obscure the substantive criticisms that challenge the fundamental foundations of post-hoc explainability methods. Challenging both the theoretical premises and practical reliability of post-hoc explainability methods, these criticisms include:

\begin{itemize}
\item \textbf{Fidelity Guarantees}: Critics such as \cite{rudin2019stop} argue that post-hoc explanation methods cannot provide genuine scientific understanding because they only approximate model behaviour rather than granting direct access to underlying mechanisms. This critique has gained formal computational support: recent complexity research demonstrates that generating explanations that are both interpretable and faithfully replicate model predictions is computationally intractable for most non-trivial models. Specifically, \cite{bassan2024local} and \cite{adolfi2025computational} prove that local and global interpretability tasks, as well as discovering concise logical circuits for inner interpretability, are typically NP-hard or worse, precluding strong fidelity guarantees except in highly restricted cases.

\item \textbf{Empirical Unreliability}: A growing body of empirical investigations \citep{hooker2019benchmark, laugel2019dangers, laugel2019issues, slack2020fooling, adebayo2022post, bilodeau2024impossibility, wei2024revisiting} has revealed pervasive limitations across the entire spectrum of post-hoc interpretability methods -- spanning local and global approaches, model-agnostic and model-specific techniques alike. For instance, \cite[pp.~7--10]{wei2024revisiting} addressed the issue of robustness in post-hoc  methods, demonstrating that many popular methods fail to provide stable explanations. Most notably, \cite{bilodeau2024impossibility} demonstrate that widely-adopted feature attribution methods (e.g., SHAP) can \textit{provably fail} to outperform random guessing when inferring model behaviour in neural networks. 

\item \textbf{Scientific Methodology Concerns}: Scientific domains remain particularly reluctant to accept post-hoc explanations as legitimate epistemic tools, with domain experts questioning their reliability for knowledge generation \citep{klein2024}. \cite{ghassemi2021false} warn that post-hoc methods create `false hope' in critical applications by appearing to provide understanding without reliable epistemic foundations. This critique fundamentally questions whether these methods can generate legitimate scientific knowledge rather than merely plausible-sounding narratives that satisfy human cognitive biases. (add Madsen et al. "empty argument") 

\item \textbf{Epistemological Imperative}: Unlike other computational research domains that can rely on predictive accuracy or measurable outcomes for validation, interpretability researchers lack the empirical validation mechanisms that typically legitimate scientific claims -- we rarely have the ground truth for benchmark evaluation in on the quality of generated explanations \citep{doshi2017towards, yang2019evaluating}, and the existence of a ground truth explanation is \textit{still} highly debated \citep{bommer2024finding}. This structural absence is compounded by the field's methodological fragmentation, where no universal consensus exists on what methodology should be applied in evaluating XAI methods \citep{ehsan2024social}, leading to repeated cycles where we take faithfulness for granted across different paradigms, later proving inadequate \citep{madsen2024interpretability}.
\end{itemize}

\textbf{Yet we argue that post-hoc interpretability methods can provide epistemically justified scientific understanding despite offering imperfect approximations of model behaviour.} Rather than demanding complete, factive explanations -- which may be unattainable for complex AI systems -- we propose that scientific knowledge emerges through (1) structured interpretation that mediates between model behaviour and phenomenal understanding, and (2) approximative explanations that acknowledge their limitations whilst maintaining empirical accountability. This position recognises that scientific progress often proceeds through strategic simplifications, and post-hoc methods, when properly validated, can serve this same epistemic function.

The remaining paper is structured as follows: Section \ref{sec:on_interp} examines the concept of interpretability in ML systems, contrasting intrinsic and post-hoc approaches. Through distinguishing between model understanding and phenomenon understanding, we establish the key theoretical relationships and assumptions that underpin our analysis. Section \ref{sec:philosophy} introduces contemporary philosophical discussions relevant to ML interpretability, making these insights accessible to the broader ML community. We examine three interconnected themes: reliability and justifiability, scientific understanding and explanation, and scientific practice in the age of AI. Section \ref{sec:defence} presents our integrated philosophical framework to defend post-hoc explainability. Through empirical case study, we also demonstrate how this framework translates theoretical principles into practical scientific discovery. Section \ref{sec:alternative} addresses and re-examines key criticisms of post-hoc methods. We offer constructive responses to concerns about approximation, fidelity, and confirmation bias. Section \ref{sec:discussion} concludes by outlining future research directions, identifying critical relationships in need of deeper theoretical examination and empirical validation. Throughout the paper, we minimise philosophical jargon where possible, with key concepts further explained in Appendix \ref{app:philosophy} for readers seeking additional clarification.

\section{On Interpretability}
\label{sec:on_interp}

Interpretability in ML systems is fundamentally pluralistic \citep{zednik2021solving}, encompassing multiple distinct concepts and serving diverse stakeholder needs. As \cite{lipton2016} and \cite{beisbart2022philosophy} emphasise, the very notion of interpretability shifts and adapts based on stakeholder expectations, posed questions, and the explanatory virtues deemed most relevant within specific contexts.

This plurality in interpretation manifests most clearly in the distinct approaches taken by different stakeholders. Computer scientists typically gravitate towards what we might term \textit{model understanding} -- a focus on comprehending the properties of the model itself, including its input-output mappings, learned representations, and decision boundaries (e.g., seeking to understand model grokking through mechanistic interpretability \citep{nanda2023progress}). In contrast, domain experts in scientific fields often seek to understand how model outputs inform or align with real-world phenomena -- what we might term \textit{phenomenon understanding}. This distinction between model and phenomenon understanding is particularly salient when examining the scientific applications of ML interpretability in detail.

For scientific applications, which are the focus of this position, interpretability encompasses our epistemic capacity to understand and articulate how AI systems generate insights about natural phenomena in ways that advance scientific understanding. This relationship begins with a true function $f(X)$ representing the natural phenomenon of interest, where the model learns an approximation $h^*(X)$ within a hypothesis space $\mathcal{H}$ that represents the set of possible functions. 

\begin{figure}[htbp]
\centering
\begin{subfigure}[b]{0.48\textwidth}
    \centering
    \includegraphics[width=\linewidth]{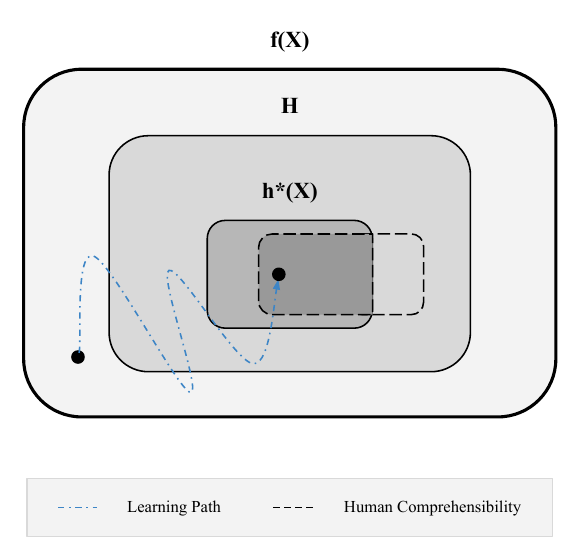}
    \caption{Intrinsically interpretable models constrain the hypothesis space $\mathcal{H}$ to human-comprehensible function classes (e.g., linear models, decision trees). The learned model $h^*(X)$ approximates the phenomenon $f(X)$ whilst remaining directly interpretable by design.}
    \label{fig:intrinsic}
\end{subfigure}
\hfill
\begin{subfigure}[b]{0.48\textwidth}
    \centering
    \includegraphics[width=\linewidth]{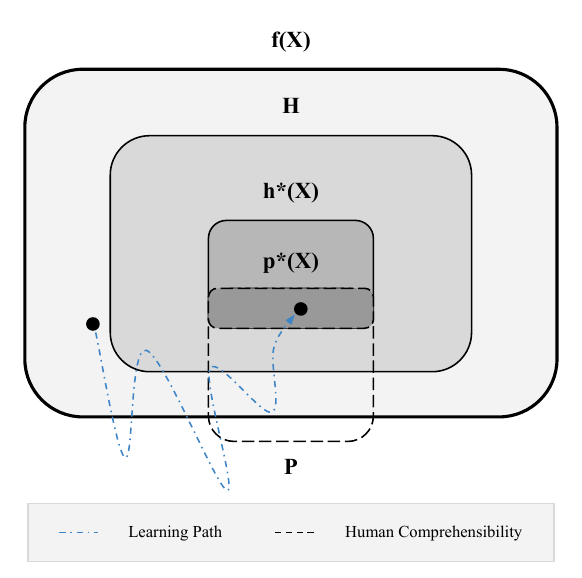}
    \caption{Post-hoc explainability allows models to be learned from an unconstrained hypothesis space $\mathcal{H}$, potentially capturing complex patterns through $h^*(X)$. A separate explanation function $p^*(X)$ is then derived (e.g., via LIME, SHAP) to approximate and interpret $h^*(X)$'s behaviour in human-understandable terms.}
    \label{fig:posthoc}
\end{subfigure}
\caption{Comparison of intrinsic and post-hoc interpretability models.}
\label{fig:interp-comparison}
\end{figure}

This broader conception is particularly important when considering post-hoc methods, whose criticisms often stem from evaluating them solely on their ability to faithfully explain a model's learnt function ($h^*(X)$) through some interpretable approximation ($p^*(X)$). However, \textit{in scientific contexts, the key relationship is not just between $h^*(X)$ and $p^*(X)$, but how interpretation relates to the underlying natural phenomenon $f(X)$ being studied}. This relationship manifests differently between intrinsically interpretable (\ref{fig:intrinsic}) and post-hoc approaches (\ref{fig:posthoc}) (Appendix \ref{app:compare}. Whilst intrinsically interpretable models maintain direct human comprehensibility by constraining $h^*(X)$, post-hoc methods introduce an additional function $p^*(X)$ learnt from a space of possible explanations $\mathcal{P}$ to approximate $h^*(X)$'s behaviour in an interpretable way. These distinct approaches illustrate the fundamental tension between model capability and human comprehensibility in scientific ML applications, where post-hoc methods potentially offer the ability to capture more complex relationships between the model and the underlying phenomenon, albeit through an additional layer of interpretation.

These relationships encompass multiple interconnected aspects: the theoretical capacity of the hypothesis space to capture the true phenomenon ($f(X)$ and $\mathcal{H}$), the representational accuracy of training data ($f(X)$ and $X$), the model selection process ($\mathcal{H}$ and $h^*(X)$), the explanation generation mechanism ($\mathcal{P}$ and $p^*(X)$), the correspondence between model and explanation spaces ($\mathcal{H}$ and $\mathcal{P}$), and the properties of the post-hoc explanations themselves ($p^*(X)$).

However, this position paper will primarily address the relationship between phenomenon and explanation ($f(X)$ and $p^*(X)$). Whilst other relationships have been extensively studied within the ML community -- such as $f(X)$ to $\mathcal{H}$ (e.g., Vapnik-Chervonenkis dimension suggesting that the capacity of the hypothesis space increases with both network width and depth \citep{bartlett2003vapnik}) and $\mathcal{H}$ to $h^*(X)$ (e.g., Structural Risk Minimisation demonstrating near-optimal learning guarantees \citep{stamouli2024structural}) -- these discussions predominantly address engineering considerations rather than epistemological foundations. 

This position aims to bring attention to the $f(X)$ to $p^*(X)$ relationship for three reasons. First, as AI models grow in complexity and opacity, $p^*(X)$ serves as our primary epistemic window into understanding complex phenomena $f(X)$ -- a role that requires timely attention as model complexity grows. Second, this relationship bridges fundamental gaps in scientific AI, such as, between black-box models and human comprehension, between computational capabilities and scientific knowledge generation, and between theoretical frameworks and their practical applications in scientific discovery. And lastly, given the growing prevalence of post-hoc methods in studying complex natural phenomena, we require a philosophical framework to validate these methods as legitimate tools for scientific knowledge production.

To further contextualise our discussion, we establish several key assumptions, beyond the fundamental premise that $h^*(X)$ effectively approximates $f(X)$. These assumptions include: 

1. \textit{Accessibility of AI Systems}: When discussing post-hoc explanations, we concentrate on open and accessible black-box algorithms, rather than proprietary systems. The primary challenge in understanding these algorithms stems from their inherent complexity, rather than a complete lack of knowledge.

2. \textit{Scientific AI Models}: Our analysis centres on supervised learning models designed to aid in science or knowledge discovery, such as predictive models in scientific research. We deliberately exclude discussion of interpretability in generative models, as they present distinct challenges beyond our current scope.

3. \textit{Imperfect but Justified Approximations}: We assume that task-specific and properly validated post-hoc methods, when embedded within structured scientific practices, can provide empirically testable insights that advance scientific understanding. Whilst these methods may offer imperfect approximations of model behaviour, their epistemic value emerges not from perfect fidelity alone, but from their ability to generate falsifiable hypotheses, guide empirical investigations, and integrate with domain knowledge through systematic validation processes.

Having established these assumptions and motivations, we now turn to examine the philosophical foundations that underpin post-hoc explainability in scientific ML. This examination is particularly timely as the ML community increasingly grapples with questions that extend beyond purely technical considerations -- questions that philosophy of science has wrestled with for centuries. By bridging these disciplines, we can leverage established philosophical frameworks to address emerging challenges in AI-driven scientific discovery, while simultaneously enriching both fields through their intersection. Moreover, as ML increasingly participate in scientific knowledge production, understanding their epistemological status becomes not merely an academic exercise, but a practical necessity for ensuring the validity and reliability of AI-generated scientific insights.

\section{Philosophical Foundations} 
\label{sec:philosophy}

\subsection{On Reliability and Justifiability}

Traditional epistemology distinguishes between two forms of justification \cite{pappas2005internalist}: internalist approaches (which require that we can consciously access and articulate our reasons for believing something) and externalist approaches (which accept that reliable processes can justify beliefs even if we cannot fully explain how they work). Whilst intrinsic interpretability aligns with internalist approaches by emphasising direct access to reasoning processes, post-hoc methods align more naturally with externalist frameworks. 

This distinction becomes particularly salient in the context of ML systems, where the inherent methodological opacity (inability to trace computational steps) and epistemic opacity (inability to understand what knowledge is encoded) of Deep Neural Networks (DNN) \cite{humphreys2009philosophical} challenges traditional notions of justification. This opacity manifests in two possible ways: through the algorithmic complexity that makes comprehensive understanding impossible, and through cognitive limitations that hinder human interpretation of the system's behaviour \cite{duran2023machine}.

Dur{\'a}n's [\citeyear{duran2023machine}] computational reliabilism (CR) addresses this challenge by proposing a framework that shifts focus from complete transparency to reliable belief-forming methods. CR delineates three categories of reliability indicators: (1) Technical Robustness of Algorithms, encompassing design, implementation, and maintenance factors; (2) Computer-based Scientific Practice, involving algorithmic implementation of scientific theories and expert assessment; and (3) Social Construction of Reliability, addressing the socially mediated processes of acceptance across scientific communities.

This framework has significant implications for post-hoc explainability methods in two ways. First, it suggests that scientific justification can emerge from the systematic validation of these methods through multiple reliability indicators, rather than requiring complete mechanistic understanding. Second, it provides a philosophical foundation for evaluating post-hoc methods based on their demonstrated reliability in connecting model behaviour to scientific phenomena, rather than their ability to provide complete transparency. This perspective reframes the epistemic status of post-hoc methods from approximate tools to potentially legitimate sources of scientific knowledge, provided they meet rigorous standards of reliability across technical, practical, and social dimensions.

\subsection{On Scientific Understanding and Explanation}

According to \cite{sullivan2022understanding}, the relationship between explanation and understanding in complex models hinges critically on the concept of ``link uncertainty" -- the gap between a model's theoretical predictions and empirical reality. Sullivan argues that while models can be epistemically opaque (meaning their internal workings are not fully transparent), this opacity does not necessarily prevent them from providing genuine understanding, provided there is sufficient empirical evidence connecting the model to real-world phenomena. In other words, we do not necessarily need to fully understand how a model works internally; what matters more is understanding how the model connects to the real-world system it is studying. 

Sullivan identifies three distinct types of explanatory questions we can ask about models: \textit{how} the model itself works, \textit{how-possibly} questions about potential mechanisms, and \textit{why/how-actually} questions about real-world phenomena. The decisive factor in moving from how-possibly to how-actually explanations is reducing link uncertainty through scientific evidence. This evidence must connect the model's theoretical insights to actual causal mechanisms in the target phenomenon. Importantly, Sullivan argues that understanding does not necessarily require complete knowledge of how a model works internally. Instead, what matters is the strength -- whether it be the amount, kind or quality -- of scientific and empirical evidence connecting the model's predictions or insights to real-world phenomena.

This perspective gains additional significance through Beisbart and R{\"a}z's [\citeyear{beisbart2022philosophy}] analysis of the factivity dilemma -- a fundamental tension between accuracy and comprehensibility in understanding DNNs. The principle of factivity (the requirement that explanations must be entirely true) demands that explanations be grounded in facts, yet modern DNNs have become so complex that we can only comprehend them through simplifications and idealisations. This creates an apparent paradox: explanations must either sacrifice accuracy for comprehensibility or maintain accuracy at the cost of being unusable.

This tension reflects a fundamental debate in philosophy of science about the relationship between explanation and understanding. Traditional accounts of scientific explanation, such as the Deductive-Nomological model, require factivity—that all elements of an explanation must be true. Yet scientific practice routinely employs strategic simplifications that sacrifice complete accuracy for tractable insights. This has led to two opposing views: factivists maintain that genuine understanding requires completely true explanations, treating simplifications as merely useful tools, whilst non-factivists argue that approximate models can provide legitimate understanding despite their imperfections. At the heart of this debate lies a crucial question: must understanding necessarily require truth, or can it emerge through useful approximations? When we refer to the ``epistemic status'' of post-hoc methods, we are asking precisely this -- whether the knowledge claims they generate can be justified and relied upon, even when they offer only approximations rather than complete truth. Importantly, this suggests a distinction between interpretability, which aims for factual explanations of model behaviour, and scientific understanding of phenomena through post-hoc methods, which may not require the same level of factivity. This distinction offers a practical resolution: whilst complete explanations remain an important goal, we can achieve meaningful scientific understanding through carefully bounded approximations.

\subsection{On Scientific Practice}

The traditional view of ML as fundamentally distinct from conventional scientific methods -- what scholars term the ``distinctness claim" -- has been challenged by Andrews’ [\citeyear{andrews2023devil}] analysis of theory-ladenness (how our theoretical assumptions inevitably shape what we observe and measure) in ML. The distinctness claim, suggesting that ML methods operate without prior theoretical assumptions, fundamentally misunderstands the nature of scientific data and practice. As Andrews demonstrates, even the most basic experimental designs reveal how data and scientific practice are inherently theory-laden, from the very act of investigation to the choice of measurement parameters.

\cite{freiesleben2024scientific} further advance this understanding by proposing ``holistic representationality" (HR) -- where we interpret the model's overall behaviour -- as an alternative to traditional ``elementwise representationality" (ER) -- where each model component directly represents something meaningful. Traditional scientific models followed what the authors call ER, where each model component -- whether parameters, variables, or relationships -- directly represented something meaningful about the phenomenon being studied. For instance, in a simple physics model, mass and velocity parameters directly correspond to physical properties. However, modern ML models, particularly DNNs, do not offer this kind of straightforward interpretation -- their individual components (e.g., network weights) do not map clearly to real-world phenomena.

Rather than viewing this as a limitation, the authors proposed HR. Instead of trying to interpret individual components, they suggest analysing the model's behaviour as a whole through what they call ``property descriptors" (e.g., cPDP, cFI, SAGE, and PRIM for global property, while ICE, cSV, ICI and Counterfactuals for local property; see pp.21-25 for further details). This approach aligns with recent findings from \cite{bilodeau2024impossibility}, demonstrating that generic feature attribution methods can be unreliable for inferring model behaviour, but task-specific approaches can dramatically improve interpretability. While Freiesleben et al. provide a theoretical framework, Bilodeau et al. offer practical evidence of its importance, showing how domain-specific interpretability methods (e.g., perturbation) can be more reliable than general-purpose approaches like SHAP or Integrated Gradients.

This modern understanding reveals ML not as distant break from traditional scientific methods, but as a new set of tools requiring theoretical understanding and methodological rigour. The framework demonstrates how post-hoc methods, when properly constructed and validated, can serve as legitimate tools for scientific inquiry. This approach maintains scientific rigour whilst acknowledging the unique characteristics of modern ML systems, suggesting a path forward that neither overstates ML's distinctness nor understates its methodological challenges.

\section{Defending Post-hoc Explainability for Scientific ML}
\label{sec:defence}

The philosophical foundations discussed above together suggest a more nuanced understanding of post-hoc interpretability methods in scientific ML. Building on these perspectives, we now develop a comprehensive framework that synthesises these insights and demonstrates their practical implications. We first present an integrated philosophical framework that establishes the epistemic validity (knowledge-generating legitimacy) of post-hoc methods in scientific inquiry, followed by a real-world case study that illustrates these principles in practice.

\subsection{An Integrated Framework}

The epistemological relationship between ML models and scientific understanding requires a synthesis of multiple philosophical frameworks to justify post-hoc interpretability methods in scientific inquiry. At its foundation lies Freiesleben et al.'s [\citeyear{freiesleben2024scientific}] distinction between elementwise representationality (ER) and holistic representationality (HR). While traditional scientific models follow ER, where each model component directly represents a physical property, modern ML models necessitate HR, where understanding emerges from analysing system-level behaviours and patterns.

This shift from ER to HR gains deeper significance when considered alongside three complementary philosophical perspectives. Andrews' [\citeyear{andrews2023devil}] theory-ladenness reveals how even our choice of interpretability methods reflects theoretical assumptions about both phenomena and models. Beisbart and R{\"a}z’s [\citeyear{beisbart2022philosophy}] factivity dilemma highlights the inherent tension between accuracy and comprehensibility in complex models. Sullivan's [\citeyear{sullivan2022understanding}] link uncertainty framework demonstrates how scientific understanding can emerge despite incomplete mechanistic knowledge.

The HR approach offers a practical resolution to these philosophical challenges. Rather than pursuing perfect component-level fidelity, it embraces holistic interpretation while maintaining scientific rigour. This aligns with empirical evidence that task-specific interpretability methods outperform general-purpose approaches \cite{bilodeau2024impossibility}, suggesting that scientific understanding requires carefully tailored methods for specific research contexts. Through this lens, post-hoc methods serve not to eliminate uncertainty, but to manage it through rigorous, testable connections between model behaviour and phenomenal understanding \cite{popper2005logic}.

Drawing on pragmatist philosophy \cite{putnam1995pragmatism}, which values practical consequences over abstract truth, we propose that post-hoc methods' epistemic value lies in their role as mediators between ML systems and human scientific understanding. This mediated nature of understanding, combined with the recognition that science often advances through strategic approximations, leads us to two key principles: \textit{mediated understanding} and \textit{bounded factivity}. \textbf{Post-hoc interpretability methods thus serve as epistemological interfaces that actively participate in knowledge falsification and expansion, capable of generating new scientific insights when maintaining scientific rigour}.

\subsubsection{Mediated Understanding}

Scientific understanding through machine learning emerges not through direct model interpretation, but through a complex process of mediated interaction. The concept of mediated understanding describes how scientific knowledge emerges through the structured interaction between four key elements: model behaviour, interpretability methods, domain knowledge, and empirical validation. This principle recognises that scientific understanding through ML is inherently mediated -- where direct access to model mechanics is neither necessary nor sufficient for scientific insight \cite{sullivan2022understanding, beisbart2022philosophy}.

The epistemic validity of post-hoc methods stems from their role as mediators in a bidirectional knowledge-creation process. In one direction (Model → Phenomenon), interpretability methods reveal patterns in model behaviour, which then tentatively suggest hypotheses about phenomena. These hypotheses, when tested empirically, provide new phenomenal understanding. In the other direction (Phenomenon → Model), domain knowledge guides the selection and refinement of interpretability methods, whilst empirical validation helps refine our interpretive approaches and identify relevant model behaviours for investigation. This bidirectional mediation provides epistemic justification because it ensures that interpretability methods are not merely describing model behaviour, but are actively participating in a cycle of hypothesis generation, empirical validation, and knowledge refinement -- the very essence of scientific inquiry.

\subsubsection{Bounded Factivity}

Building on our discussion of mediated understanding, we now turn to bounded factivity, which helps resolve fundamental tensions between the approximative nature of post-hoc methods and their epistemic value for scientific understanding. Rather than demanding complete factivity -- perfect correspondence between interpretation and model mechanics -- bounded factivity acknowledges truth within explicitly acknowledged limits and simplifications. This aligns with traditional scientific practice, where scientists routinely use simplified models that deliberately deviate from reality to gain understanding of complex phenomena \cite{beisbart2022philosophy, sullivan2022understanding}.

The recognition of strategic simplification's role in science helps reconceptualise the epistemic status of post-hoc interpretability methods. Recent empirical work by \cite{bilodeau2024impossibility}] demonstrates that whilst many general-purpose post-hoc methods may be unreliable, carefully designed, task-specific approaches can provide reliable insights within bounded contexts. By aligning interpretability methods with specific scientific goals \cite{freiesleben2024scientific} and validating them through systematic empirical testing \cite{sullivan2022understanding}, we can achieve meaningful understanding within acknowledged bounds, just as traditional scientific models advance understanding despite their simplifications.

\subsubsection{Structured Process of Scientific Discovery}

\begin{figure*}[htbp]
\centering
\includegraphics[width=\textwidth]{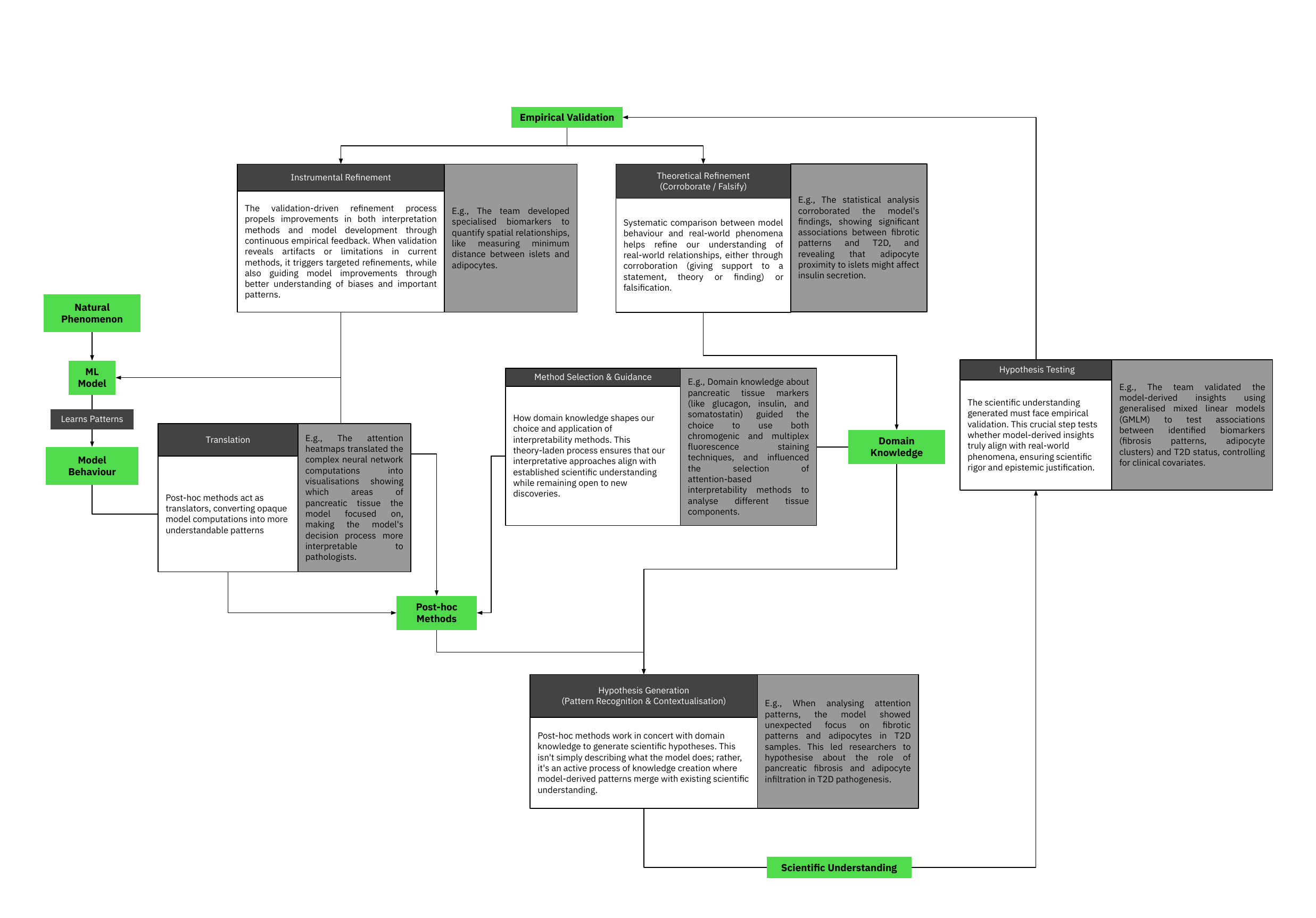}
\caption{Framework for Scientific Knowledge Generation through Post-hoc Methods. The diagram illustrates the cyclic process of generating scientific understanding from ML models through post-hoc interpretability methods. Each component shows both the theoretical principle (dark grey boxes) and its practical application in T2D research (light grey boxes). The process begins with translation of model behaviour, progresses through method selection and hypothesis generation, and culminates in empirical validation through instrumental and theoretical refinement. Green elements represent the key stages in the knowledge generation pipeline. Arrows indicate the flow and interactions between components, demonstrating how post-hoc methods mediate between model behaviour and scientific understanding.}
\label{fig:casestudy}
\end{figure*}

The philosophical framework we have established can be operationalised into a structured process for generating scientific knowledge through post-hoc methods (\ref{fig:casestudy}, \ref{app}). This process, illustrated through a cyclic interaction between model behaviour, interpretability methods, and domain knowledge, demonstrates how post-hoc methods can contribute to scientific understanding whilst maintaining epistemic rigour. The diagram pairs each theoretical component with a concrete example from type 2 diabetes research (detailed in the next subsection), providing a practical roadmap for how post-hoc interpretability methods can generate legitimate scientific knowledge.

\subsection{Bridging Philosophy and Practice: A Case Study}

The recent work by \cite{klein2024} on explainable AI-based analysis of pancreatic tissue in type 2 diabetes (T2D) provides a compelling demonstration of our philosophical framework in action. Their study exemplifies how post-hoc explainability methods can generate legitimate scientific knowledge through systematic application of mediated understanding and bounded factivity principles.

The study demonstrates mediated understanding through carefully orchestrated interactions between model behaviour, interpretability methods, and domain expertise. The translation phase employed multiple XAI techniques -- attention mechanisms and attribution methods —- to convert complex model computations into interpretable visualisations. This translation was inherently theory-laden: the selection of six immunostaining markers (insulin, glucagon, somatostatin, PECAM1, perilipin 1, and tubulin beta 3) reflected theoretical understanding whilst remaining open to novel discoveries. Remarkably, their best-performing model focused on islet $\alpha$- and $\delta$-cells along with neuronal structures rather than $\beta$-cells, challenging traditional assumptions about T2D pathophysiology. This unexpected finding emerged through the interaction between model behaviour and domain expertise, demonstrating how post-hoc methods can reveal insights that transcend existing theoretical frameworks.

The bounded factivity principle manifests powerfully in their validation approach. Rather than claiming complete mechanistic understanding, the researchers demonstrated scientific rigour -- the strict application of the scientific method to experimental design and interpretation \citep{klein2024, weber2024xai} -- by explicitly bounding their claims within testable hypotheses. They developed targeted biomarkers to quantify specific observations: adipocyte-to-islet proximity (measured in micrometres), fibrosis density patterns, and morphological features. These biomarkers underwent systematic validation through generalised mixed linear models (GMLM), controlling for clinical covariates (age, sex, BMI) and random effects (cohort, staining method). Notably, they achieved statistical significance for key associations, \textit{whilst} acknowledging limitations such as segmentation challenges.

\cite{klein2024} compellingly demonstrated how post-hoc explainability methods advance scientific understanding beyond mere model interpretation. Their discovery that islets in T2D patients are significantly closer to adipocytes suggests paracrine effects impacting islet function -- a hypothesis that emerged directly from XAI-guided analysis. The unexpected finding that $\alpha$-cells, $\delta$-cells, and neuronal markers outperformed $\beta$-cell markers in prediction challenges the $\beta$-cell-centric view of T2D and suggests more complex pathophysiology involving islet innervation. These insights were not merely descriptions of model behaviour but genuine contributions to understanding T2D pathogenesis, validated through statistical analysis and grounded in biological plausibility. The study thus exemplifies how post-hoc methods, when embedded within structured scientific practices and constrained by bounded factivity, serve as legitimate tools for scientific discovery.

\section{Re-assessing Alternative Views}
\label{sec:alternative}


\subsection{Approximation and Fidelity}

\cite{rudin2019stop} and \cite{ghassemi2021false} argue that post-hoc explanations are problematic due to their approximative nature. This critique necessitates careful examination of distinct but related concerns: the factivity of explanations and the nature of understanding in scientific practice. The dilemma presents itself thus, \textit{completely accurate explanations of complex ML models would merely duplicate their opacity, whilst simplified explanations necessarily introduce some degree of falsehood}. This apparent tension can be productively addressed through the lens of non-factive understanding in science \cite{beisbart2022philosophy}.

Interpretability exists on a spectrum, with increased epistemic value and practical utility correlating with higher degrees of interpretability. This approach aligns with the concept of \textit{verisimilitude} (``closeness to truth''), where approximations to truth, though imperfect, retain epistemic worth \cite{oddie2001}. Although post-hoc explanations lack performance guarantees and do not fully capture model behaviour, this limitation need not compromise their epistemological value if we maintain awareness of departure \cite{kvanvig2009responses} -- conscious recognition of where and how our explanations diverge from ground truth. Many scientific and analytical tools rely on strategic idealisations that, despite their non-factive nature, provide valuable insights and practical utility. The key is maintaining empirical accountability through testable predictions, situating approximations within relevant theoretical frameworks, and providing clear scope conditions for the validity of interpretations.

Just as scientific models generally involve idealisations that technically violate factivity without compromising their utility for understanding, post-hoc explanations can provide genuine scientific insight even while containing strategic simplifications. This suggests that the key to maintaining scientific rigour lies not in perfect factivity, but in transparent acknowledgment of simplifications coupled with continuous refinement through empirical validation. Both intrinsically interpretable models and post-hoc explanations are simplifications of the complex systems they represent. Accepting an intrinsically interpretable model as `understandable' and having some fidelity to the real world is philosophically analogous to accepting a post-hoc explanation that is `understandable' and has some fidelity to the original model. The fidelity between complex systems (real world or AI) and any model (intrinsic or post-hoc) is inherently imperfect, yet this imperfection does not negate their scientific value when properly bounded and validated.

\subsection{Faithful Explanation and Confirmation Bias}

Rudin's [\citeyear{rudin2019stop}] critique of post-hoc methods focuses on two potential pitfalls: incomplete (local) explanations and unjustifiable explanations. However, local explanations could serve as distinct epistemic tools that offer granular insights into model behaviour. These local insights can generate testable hypotheses about both model behaviour and phenomenal relationships, identify edge cases that reveal important patterns, and expose nuances that global explanations might miss. Local explanations may complement rather than compete with global understanding. Regarding unjustifiable explanations, our framework suggests that even apparently problematic model behaviours -- such as scientifically unsound judgments or confounding variables -- can advance scientific understanding when properly interpreted. We recognise that identifying flaws in model reasoning contributes valuable knowledge about both model limitations and phenomenal complexity. This aligns with how sciences historically progress through understanding both positive and negative results.

\cite{ghassemi2021false} raise a complementary concern about confirmation bias in interpreting post-hoc explanations, suggesting that humans might draw overconfident conclusions from potentially unreliable interpretations. This "interpretability gap" could potentially foster false confidence in the model's reliability or fairness. The limitations \citeauthor{ghassemi2021false} describe are not unique to AI explanations but are inherent in complex judgements, whether human or artificial.  Human experts, like AI systems, can fall prey to confirmation bias, potentially leading to overconfidence in their interpretations or explanations. Following this reasoning to its logical conclusion, one might argue that we should be equally sceptical of human expert explanations as we are of AI-generated ones. Taken to an extreme, this line of thinking could lead to an argument for minimising reliance on expert explanations altogether, whether human or AI-generated. Instead, one might advocate for sole reliance on predefined, explicit "if-then" rules, aiming to eliminate the subjectivity and potential biases inherent in both human and AI interpretations. Yet, this conclusion overlooks the value of both human and AI-generated post-hoc explanations.

Rather than suggesting we should abandon post-hoc explanations in favour of purely rule-based approaches, our framework advocates for their refinement and systematic validation. Just as sciences have developed methods for managing human cognitive biases while preserving the value of expert insight, we can develop approaches to post-hoc interpretation that acknowledge limitations while maximising epistemic value through careful bounds and empirical validation.

\section{Discussion}
\label{sec:discussion}

This position paper has several limitations that warrant acknowledgement. First, our philosophical framework relies primarily on a single empirical case study (Klein et al., 2024) to demonstrate practical application. Whilst this study provides some evidence, additional examples across diverse scientific domains would strengthen our argument. Second, we focus exclusively on supervised learning for scientific discovery, excluding important ML paradigms such as unsupervised learning, reinforcement learning, and generative models, which may require different epistemological considerations. Third, we deliberately focus on epistemic rather than normative considerations, leaving important questions about the ethical obligations and social responsibilities of XAI unexplored (see Appendices \ref{app:vreden} and \ref{app:lazar} for relevant normative discussions). Finally, whilst we introduce the concept of "bounded factivity," we remain at the philosophical level without operationalising it into measurable criteria. Though \citet{wei2024revisiting} demonstrate fine-grained metrics for evaluating post-hoc method robustness, translating such approaches to scientific contexts remains unclear (add Friesleban). What constitutes sufficient "bounds" for scientific practice? How do we measure faithfulness when the goal is phenomenal understanding rather than model fidelity? Our focus was establishing how post-hoc methods can be philosophically justified within scientific practices; developing empirical measures and concrete desiderata informed by philosophy of science represents an important but separate endeavour for future work.

Attempts to interpret AI models, particularly through post-hoc methods, bear a meaningful resemblance to the way human experts articulate their intuitive judgements. Much like post-hoc rationalisations offered by human specialists -- who often cannot access the full breadth of their internalised cognitive processes—these AI explanations seek to render comprehensible the dense, layered computations embedded in modern models \citep{newell1972human, Simon1973, gobet2004chunks}. This parallel highlights not only the approximative nature of both forms of explanation but also a critical distinction: unlike the human mind, AI systems permit systematic interrogation of their internal states. This accessibility allows for more transparent bounds on factivity and the possibility of empirical validation—advantages that position post-hoc explainability as a potentially more rigorous and inspectable form of interpretation than human expertise affords. In this view, post-hoc methods are not peripheral conveniences but central instruments in the pursuit of scientific insight from ML systems. While their approximative nature imposes limits, these limits are not disqualifying -- provided they are acknowledged, bounded, and subjected to empirical scrutiny. Post-hoc explainability, when properly constrained and philosophically grounded, can be an effective conduit for phenomenal understanding within scientific practice.

\bibliographystyle{plainnat}
\bibliography{neurips25}

\newpage
\appendix

\section{Conceptual Comparison of Intrinsic and Post-hoc Interpretability Models}
\label{app:compare}

\newpage
\section{Framework for Scientific Knowledge Generation through Post-hoc Methods}
\label{app:framework}

\newpage
\section{Relevant Philosophical discussions on AI and Interpretability}
\label{app:philosophy}

\subsection{\citeauthor{sullivan2022understanding}’s (\citeyear{sullivan2022understanding}) Link Uncertainty}
\label{app:sullivan}

According to Sullivan, the relationship between explanation and understanding in complex models hinges critically on the concept of "link uncertainty" -- the gap between a model's theoretical predictions and empirical reality. Sullivan argues that while models can be epistemically opaque (meaning their internal workings are not fully transparent), this opacity does not necessarily prevent them from providing genuine understanding, provided there is sufficient empirical evidence connecting the model to real-world phenomena. In other words, we do not necessarily need to fully understand how a model works internally; what matters more is understanding how the model connects to the real-world system it is studying. 

Sullivan identifies three distinct types of explanatory questions we can ask about models: how the model itself works, how-possibly questions about potential mechanisms, and why/how-actually questions about real-world phenomena. How-possibly explanations demonstrate potential mechanisms or causes, showing how something could theoretically occur. However, these fall short of explaining how things actually work in reality. Using Schelling's segregation model as an example, Sullivan shows that while the model can demonstrate how segregation could emerge from individual preferences, it only provides genuine understanding if there is empirical evidence showing these mechanisms actually operate in real-world segregation patterns.

The decisive factor in moving from how-possibly to how-actually explanations is reducing link uncertainty through scientific evidence. This evidence must connect the model's theoretical insights to actual causal mechanisms in the target phenomenon. Importantly, Sullivan argues that understanding does not necessarily require complete knowledge of how a model works internally. Instead, what matters is the strength -- whether it be the amount, kind or quality -- of scientific and empirical evidence connecting the model's predictions or insights to real-world phenomena.

\subsection{\citeauthor{andrews2023devil}' (\citeyear{andrews2023devil}) Theory-ladenness of Machine Learning}
\label{app:andrews}

The debate over Machine Learning's impact on science has generated what scholars call the "distinctness claim". The claim's core argument -- articulated by several philosophers like Boge, Sreckovic et al., and Boon -- is that ML, particularly deep learning, represents a fundamental departure from traditional scientific methods. They primarily base this on two key distinctions: (1) ML methods are supposedly "theory-free" or "theory-agnostic", operating without prior theoretical assumptions or conceptualisations of target phenomena, and (2) ML models prioritise prediction over explanation and understanding, making them epistemically opaque in novel ways. This perspective has gained significant traction not only in philosophical discourse but also among scientists and engineers who view ML as fundamentally different from traditional scientific approaches.

Extending Leonelli, Andrews fundamentally challenges this perspective with the theory-laden nature of scientific data and practice:

\begin{quotation}
Even the most simplistic of experimental designs reveals the nature and extent to which data, and scientific practice at large, are 'theory-laden.' The very act of investigation involves commitment to the existence and in-principle measurability of some phenomenon and, if we are making measurements and performing quantitative analyses thereon, commitment to its quantitative nature... 
\\
Measurement cannot be total, and therefore there is always a commitment as to what to look at experimentally and what to exclude. There is always a commitment to the appropriate level of abstraction at which to study the phenomenon in play in terms of such things as instrument settings like degree of magnification or periodicity of sampling. The very design of our instruments of measure and their calibration includes various commitments to the nature of the worldly phenomena under investigation. \citep[pp.~6]{andrews2023devil}
\end{quotation}

This understanding of data's theory-laden nature is now widely accepted in philosophy of science. However, as Leonelli notes, unfortunate relics of this view -- viewing data as mere 'empirical input for modelling' -- remain widespread. This persistent misconception underlies many arguments about ML's theory-free nature. The reality is that all scientific data, whether used in traditional methods or ML, necessarily involves theoretical assumptions and conceptual frameworks in its collection, preparation, and interpretation. This view challenges the technological determinism implicit in many discussions of ML in science -- the belief that certain effects or limitations of ML are fixed, inevitable consequences of the technology itself. Rather than accepting current limitations as inherent features, Andrews argues we should recognise them as methodological challenges that can be addressed through improved practices and understanding.

Building on this theoretical foundation, Andrews demonstrates how the impossibility of "theory-free" learning is established by both philosophy of science and theoretical computer science's understanding of inductive generalisation. At its core, machine learning performs inductive inference - extrapolating from limited instances to general cases. Drawing on Norton's material theory of induction, Andrews notes that successful inductive inference never proceeds through universal, domain-generic formal rules, but rather requires the application of local rules warranted by empirical facts specific to each research domain. This philosophical insight finds independent confirmation in computer science through the No Free Lunch theorems, which mathematically demonstrate the impossibility of universal domain-generic inference rules. While these theorems were derived in specific formal settings, their implications for ML practice are profound: inductive inference fundamentally requires domain-specific inductive biases. This convergence of philosophical and mathematical results undermines claims about ML's theory-independence.

These insights reveal the "distinctness claim" as fundamentally misguided. Rather than representing a revolutionary break from traditional scientific methods, ML should be understood as a new set of tools whose proper application still requires theoretical understanding and methodological rigour. This perspective suggests a more nuanced approach to ML in science: one that acknowledges how theoretical considerations may enter differently in ML workflows, while recognising their essential role in ensuring sound scientific practice. Such an understanding is crucial for developing appropriate methodological standards for ML in science, rather than accepting current limitations as inevitable features of the technology.

\subsection{\citeauthor{beisbart2022philosophy}'s (\citeyear{beisbart2022philosophy}) Factivity Dilemma}
\label{app:beisbart}

The factivity dilemma in understanding Deep Neural Networks (DNNs) centers on a fundamental tension between accuracy and comprehensibility. The principle of factivity demands that explanations and understanding be grounded in facts, yet modern DNNs have become so complex that we can only comprehend them through simplifications and idealisations. As Rudin (2019) pointedly argues, a perfectly accurate explanation would simply duplicate the original model's complexity, defeating the purpose of explanation. This creates what appears to be an insurmountable challenge: explanations must either sacrifice accuracy for comprehensibility or maintain accuracy at the cost of being unusable.

This tension has deep roots in the philosophy of science, particularly in debates about the relationship between explanation and understanding. Traditional accounts of scientific explanation, such as the Deductive-Nomological model, typically require factivity - the premises in an explanation must be true. However, the requirements for scientific understanding are more nuanced. Non-factivists like Elgin argue that simplified models can provide legitimate understanding despite imperfect accuracy, while factivists such as Lawler maintain that simplifications are merely instruments toward understanding rather than constituting understanding itself. These opposing views reflect a broader debate about whether understanding necessarily requires truth or can be achieved through useful approximations.

A potential resolution emerges when we distinguish between mechanistic interpretability and scientific understanding in the context of DNNs. While mechanistic interpretability aims for factual explanations of model behaviour, scientific understanding of phenomena through post-hoc interpretative models may not require the same level of factivity. This distinction suggests that while complete, accurate explanations remain an important goal, we can develop meaningful understanding through carefully constructed simplified models. The key lies in maintaining awareness of these models' limitations while leveraging their insights - acknowledging them as useful approximations rather than complete representations of reality. This approach offers a practical way forward, recognising both the current constraints in explaining DNNs and the necessity of working with these systems, even with imperfect understanding.

\subsection{\citeauthor{freiesleben2024scientific}'s (\citeyear{freiesleben2024scientific}) Holistic Representationality} 
\label{app:freiesleben}

Freiesleben et al. address a fundamental challenge in modern scientific research: how to derive meaningful scientific insights from machine learning models that, unlike traditional scientific models, lack direct interpretability of their components. Traditional scientific models followed what the authors call "elementwise representationality" (ER), where each model component -- whether parameters, variables, or relationships -- directly represented something meaningful about the phenomenon being studied. For instance, in a simple physics model, mass and velocity parameters directly correspond to physical properties. However, modern ML models, particularly neural networks, do not offer this kind of straightforward interpretation - their individual components (like network weights) do not map clearly to real-world phenomena (e.g., see \citet{freiesleben2023artificial}).

Rather than viewing this as a limitation, the authors propose a framework based on "holistic representationality" (HR). Instead of trying to interpret individual components, they suggest analysing the model's behaviour as a whole through what they call "property descriptors" (e.g., cPDP, cFI, SAGE, and PRIM for global property, and ICE, cSV, ICI and Counterfactuals for local property, see pp.21-25 for further details). This approach aligns with recent findings from \citet{bilodeau2024impossibility}, who demonstrate that generic feature attribution methods can be unreliable for inferring model behaviour, but task-specific approaches can dramatically improve interpretability. While Freiesleben et al. provide a theoretical framework, Bilodeau et al. offer practical evidence of its importance, showing how domain-specific interpretability methods (e.g., perturbation) can be more reliable than general-purpose approaches like SHAP or Integrated Gradients.

The authors provide a systematic four-step framework for this approach (pp. 14-20): first, formalising the scientific question as a statistical query, which involves translating research questions into precise mathematical formulations; second, identifying how to answer it using the whole model through property descriptors that are continuous functions mapping from model space to answer space; third, estimating the answer using the trained model, which requires careful consideration of data distribution and model behavior; and fourth, quantifying the uncertainty in the results through both model error (difference between optimal and trained model predictions) and estimation error (uncertainty in the property descriptor estimates themselves). This framework is particularly notable for its rigorous treatment of uncertainty quantification, which is often overlooked in traditional interpretable ML approaches.

The paper demonstrates the practical applicability of this framework by showing how existing interpretable ML methods can serve as property descriptors. Using a concrete example of analysing student academic performance, they illustrate how these methods can provide scientifically meaningful insights while maintaining rigorous standards of inference. The authors emphasise that while this approach differs from traditional scientific modelling, it does not sacrifice scientific rigour -- it simply provides a different path to extracting knowledge from our models, one that's better suited to the capabilities and limitations of modern machine learning systems. This conclusion resonates with Bilodeau et al.'s findings that success in model interpretation often depends on carefully defining concrete end-tasks and developing targeted evaluation methods rather than relying on general-purpose interpretation tools.

\subsection{\citeauthor{lazar2024legitimacy}'s (\citeyear{lazar2024legitimacy}) Democratic Duties of Explanation}
\label{app:lazar}

Lazar's central contribution to AI explainability discourse stems from his recognition that computational systems, especially AI, are increasingly being used to "govern" us -- that is, to settle, implement, and enforce the norms that determine how institutions function. When computational systems are deployed by government agencies in administrative functions or by private companies to police online behaviour and determine our information access, they are effectively governing us. For such governing power to be legitimate, Lazar argues, it must be accountable to democratic oversight through public explanation to the community as a whole. Unlike approaches focused on individual rights or technical transparency, Lazar emphasises that explainability is fundamentally a democratic duty -- it is not about individual decision subjects understanding their particular outcomes, but about enabling the collective community to determine whether these computational governance systems are being used legitimately and with proper authority. Lazar argues that this collective explainability requirement has specific implications for computational governance systems: they must reveal not just their decision rules, but also demonstrate the appropriateness of their training data as evidence, the robustness of their decision-making processes, and their ability to make the right decisions for the right reasons. 

\subsection{\citeauthor{vredenburgh2022right}'s (\citeyear{vredenburgh2022right}) Informed Self-advocacy}
\label{app:vreden}

Vredenburgh's central contribution addresses the fundamental tension between algorithmic opacity and individual rights. Rather than demanding complete technical transparency of complex AI models, she argues for a claim right to explanations that can be provided post-hoc, grounded in what she calls "informed self-advocacy" -- a cluster of abilities that allows individuals to represent their interests and values to decision-makers and further those interests within institutions. This right becomes particularly crucial in institutions where algorithmic decisions significantly impact individuals' lives.

Vredenburgh argues that post-hoc explanations must take two specific forms: rule-based normative explanations (explaining why a decision was appropriate) and rule-based causal explanations (explaining how inputs relate to outputs). She advocates for "functional transparency" -- high-level explanations of how inputs relate to outputs -- rather than structural or run transparency of the underlying model (pp. 13). While acknowledging that simplified explanations of complex algorithms may be somewhat inaccurate, she argues they can still be sufficient for informed self-advocacy if properly calibrated to stakes: when decisions distribute harms or entitlements (versus benefits), there are stronger requirements for clear explanations and human expert support. This pragmatic framework shows how post-hoc explanations, even if they do not fully capture the complexity of AI systems, can satisfy legitimate needs for accountability while remaining feasibly implementable, as evidenced by existing legal requirements for explanation across various domains.

\subsection{\citeauthor{duran2023machine}'s (\citeyear{duran2023machine}) Computational Realibilism} 
\label{app:duran}

The central motivation of justification in DNNs is primarily driven by their inherent methodological and epistemic opacity \citep{humphreys2009philosophical}. This opacity manifests in two distinct yet interrelated ways \citep{duran2023machine}. First, the algorithmic complexity of DNN systems -- encompassing myriad functions, variables, decisions, and data -- renders it impossible for any individual or group to fully comprehend which elements are pivotal in generating a specific output. Second, this complexity imposes cognitive limitations on human agents, hindering our ability to derive meaningful interpretations of the algorithm and its results. Both aspects of opacity potentially undermine the justificatory basis for ascribing scientific value to DNN outputs, either due to the "black-box" nature of the system or the cognitive constraints of human interpreters.

Duran's computational reliabilism (CR) addresses this epistemic challenge by proposing a framework for justifying belief in DNN outputs if and when they are produced by reliable belief-forming methods \citep{duran2018grounds, duran2023machine, javed2023trust}. CR delineates three categories of reliability indicators: (i) Technical Robustness of Algorithms, encompassing the design, implementation, and maintenance factors that contribute to a DNN system's robustness; (ii) Computer-based Scientific Practice, which involves the algorithmic implementation of scientific theories and principles, or expert assessment within established scientific knowledge; and (iii) Social Construction of Reliability, referring to the socially mediated processes that confer acceptance of DNN and its outputs across diverse communities. At its core, CR adopts a frequentist approach, positing that beliefs formed by demonstrably reliable algorithms warrant greater justification than those produced by unreliable ones.

\end{document}